\pdfoutput=1
\documentclass[11pt]{article}

\usepackage[final]{acl}

\usepackage{times}
\usepackage{latexsym}

\usepackage[T1]{fontenc}
\usepackage[utf8]{inputenc}

\usepackage{microtype}

\usepackage{inconsolata}

\usepackage{graphicx}

\title{TableLoRA: Low-rank Adaptation on Table Structure Understanding \\ for Large Language Models}

\makeatletter
\newcommand{\printfnsymbol}[1]{%
  \textsuperscript{\@fnsymbol{#1}}%
} 
\makeatother
\author{
Xinyi He\textsuperscript{\rm 1}\thanks{\indent The contributions by Xinyi He and Yihao Liu have been conducted and completed during their internships at Microsoft.}\hspace{0.5em}
Yihao Liu\textsuperscript{\rm 2}\printfnsymbol{1}\hspace{0.5em}
Mengyu Zhou\textsuperscript{\rm 3}\thanks{\indent Corresponding author.}\hspace{0.5em} 
Yeye He \textsuperscript{\rm 3}\hspace{0.5em}
\\
\textbf{Haoyu Dong} \textsuperscript{\rm 3}\hspace{0.5em}
\textbf{Shi Han} \textsuperscript{\rm 3}\hspace{0.5em}
\textbf{Zejian Yuan}\textsuperscript{\rm 1}\hspace{0.5em}
\textbf{Dongmei Zhang}\textsuperscript{\rm 3}\hspace{0.5em} \\
\textsuperscript{\rm 1} State Key Laboratory of Human-Machine Hybrid Augmented Intelligence, \\Institute of Artificial Intelligence and Robotics, Xi'an Jiaotong University \\
 \textsuperscript{\rm 2} Peking University 
\textsuperscript{\rm 3} Microsoft\\
\texttt{\href{mailto:hxyhxy@stu.xjtu.edu.cn}{hxyhxy@stu.xjtu.edu.cn}},
\texttt{\href{mailto:haoeliu@stu.pku.edu.cn}{haoeliu@stu.pku.edu.cn}},
\texttt{\href{yuan.ze.jian@xjtu.edu.cn}{yuan.ze.jian@xjtu.edu.cn}}, \\
\texttt{\{\href{mailto:mezho@microsoft.com}{mezho}, \href{mailto:yeyehe@microsoft.com}{yeyehe}, \href{mailto:haoyu.dong@microsoft.com}{haoyu.dong}, \href{mailto:shihan@microsoft.com}{shihan}, \href{mailto:dongmeiz@microsoft.com}{dongmeiz}\}@microsoft.com}}

\usepackage{float}
\usepackage{subfig}
\usepackage{booktabs} % for professional tables
\usepackage{amsmath}
\usepackage{amsthm}
\usepackage{amsfonts}       % blackboard math symbols
\usepackage{nicefrac}       % compact symbols for 1/2, etc.
\usepackage{xspace}
\usepackage{xcolor}
\usepackage{backnaur}
\usepackage{multirow}
\usepackage{makecell}
\usepackage{appendix}
\usepackage{enumitem}
\usepackage{circledsteps}
\usepackage{color}
\usepackage{colortbl}
\usepackage{graphicx}
\usepackage{amsmath}
\usepackage{caption}  
\usepackage{listings}
\usepackage{verbatim}
\usepackage{amssymb}
\theoremstyle{definition}

\newcommand{\refequ}[1]{Equation~(\ref{#1})}
\newcommand{\reffig}[1]{Figure~\ref{#1}}
\newcommand{\refsec}[1]{\S\ref{#1}} % \textsection
\newcommand{\reftab}[1]{Table~\ref{#1}}

\def\eg{\textit{e.g.}\xspace}
\def\Eg{\textit{E.g.}\xspace}

\def\etc{\textit{etc.}\xspace}

\begin{document}
\maketitle
\begin{abstract}
Tabular data are crucial in many fields and their understanding by large language models (LLMs) under high parameter efficiency paradigm  is important.
However, directly applying parameter-efficient fine-tuning (PEFT) techniques to tabular tasks presents significant challenges, particularly in terms of better table serialization and the representation of two-dimensional structured information within a one-dimensional sequence. 
To address this, we propose TableLoRA, a module designed to improve LLMs' understanding of table structure during PEFT. It incorporates special tokens for serializing tables with special token encoder and uses 2D LoRA to encode low-rank information on cell positions. Experiments on four tabular-related datasets demonstrate that TableLoRA consistently outperforms vanilla LoRA and surpasses various table encoding methods tested in control experiments. These findings reveal that TableLoRA, as a table-specific LoRA, enhances the ability of LLMs to process tabular data effectively, especially in low-parameter settings, demonstrating its potential as a robust solution for handling table-related tasks.
\end{abstract}

\section{Introduction}

% \TODO{highlight vs finetune}
Tabular data are widely used in numerous fields, and LLMs are also widely applied to the understanding and processing of tabular data, such as Table-GPT~\cite{li2024table-gpt}, TableLLM~\cite{zhang2024tablellm}, \etc. Meanwhile, the PEFT~\cite{peft} (Parameter-Efficient Fine-Tuning) paradigm, with its advantages of high parameter efficiency, is widely used to fine-tune LLMs. 
% And to achieve an effective balance between performance and efficiency, PEFT is also applied to tabular data , for example, TableLlama~\cite{zhang2023tablellama} utilizes LongLoRA making it a large generalist model for tables.
Consequently, exploring methods to learn improved table representations under a high parameter efficiency paradigm to address table-related issues more effectively remains a critical and valuable area of research.

\begin{figure}[tb]
    \centering
    \includegraphics[width=1\linewidth]{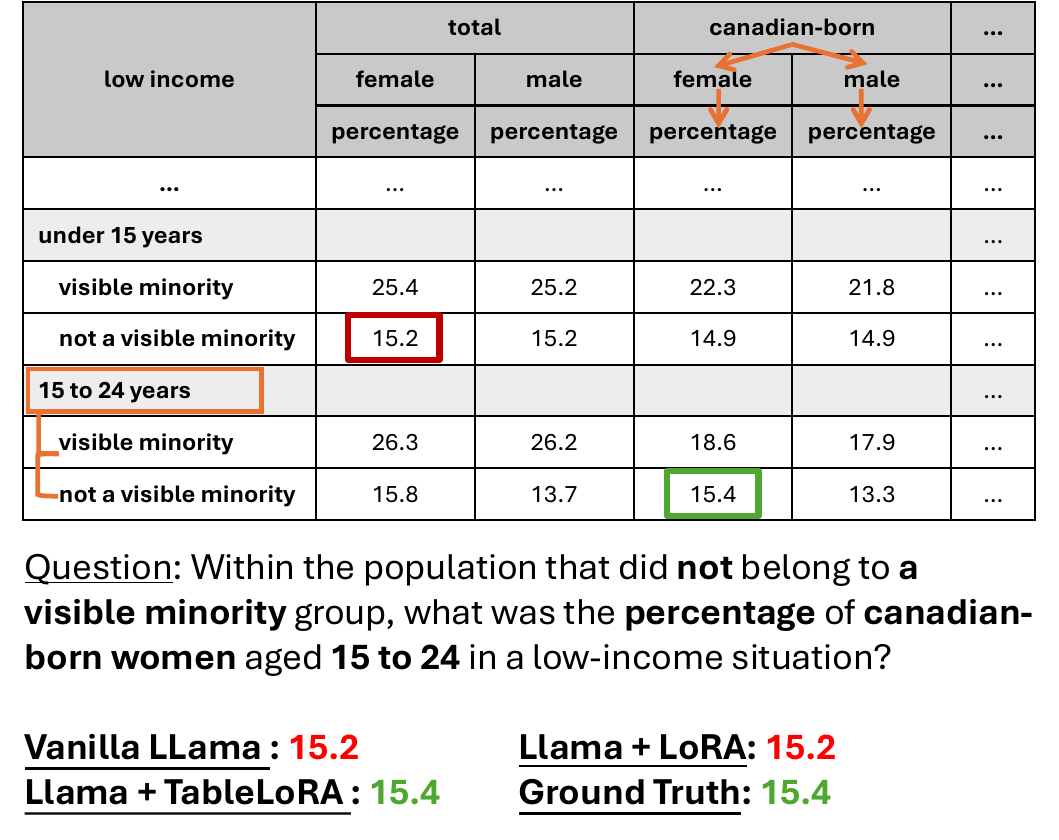}
    \caption{An Example of Common Errors when Directly Using LoRA to Finetune Llama on Tabular Tasks.}
    \label{fig:error_eg}
    \vspace{-3mm}
\end{figure}

% However, after directly using PEFT to fine-tune the model, LLMs still make naïve errors in understanding table structures. For example, in the TableQA example shown in \reffig{fig:error_eg}, it is necessary to locate rows and columns to retrieve cell values from the table. Retrieving data has already become a relatively basic task in data analysis. However, even after LoRA fine-tuning, Llama still makes mistakes, and both row and column locations are incorrect. This indicates that LoRA has limited learning capabilities for table structures. Furthermore, its understanding of tree-structured headers in tables needs improvement.

However, directly applying PEFT(\eg, Low-rank Adaptation(LoRA)) to table-related tasks reveals several critical challenges:

\textit{Challenge 1}: How to better serialize tables. 
Previous studies have shown that different methods of table serialization affect the results~\cite{sui2024tablemeetsllmlarge}. However, even with existing serialization techniques, models still struggle to accurately recognize table structures. \Eg, in the TableQA example shown in \reffig{fig:error_eg}, it needs to retrieve cell from the table. The column for retrieval can be easily identified through the same header names in the query. However, the Llama model fine-tuned with LoRA fails to recognize the cell in the same column as the header.

\textit{Challenge 2}: How to better represent two-dimensional structured information in a one-dimensional sequence. The positional information of rows and columns in a table is crucial for understanding the table structure, row-column correspondence, and so on. \Eg, to comprehend the hierarchical left header in \reffig{fig:error_eg}, one must recognize that cells like "15 to 24 years" are in the leftmost column and identify their corresponding row and column content. However, during LoRA fine-tuning, positional information is not explicitly learned and is only implicitly computed through attention, leading to the query in \reffig{fig:error_eg} being unable to accurately locate the rows to retrieve through the hierarchical structure of the left header.

% To help LLMs understand table structure during parameter-efficient fine-tuning, we propose \textbf{TableLoRA}\footnote{The code will be publicly available upon acceptance.} as a compatible module for PEFT of LLMs. Existing tabular LLMs attempt to learn table structure relationships through extra training using attention mechanisms. We directly inform the model of these structural relationships through model design, which is divided into two parts: 1) Adding special tokens for serializing tables into the prompt. 2) Row/column 2D LoRA for encoding low-rank information about cell position.

To promote the recognition and understanding of table structures during tabular tasks and enhance the ability of Large Language Models (LLMs) to process tables within PEFT, we propose TableLoRA\footnote{The code will be open-sourced on https://github.com/microsoft/TableLoRA.}, a module compatible with the PEFT framework for LLMs. Existing tabular LLMs attempt to learn table structure relationships using attention mechanisms through additional training. In contrast, we directly inform the model of these relationships through our design, which consists of two key components:

To address \textit{Challenge 1}, the \textbf{Special Tokens Encoder} enhances tabular data representation during fine-tuning. The challenge lies in effectively learning and incorporating the special tokens $\texttt{[tab]}$, $\texttt{[row]}$, and $\texttt{[cell]}$, which replace traditional markdown or HTML formats and provide a structured tabular representation for improved model processing. We use a fine-tuning method inspired by p-tuning to ensure effective gradient propagation to special token embeddings, enhancing the model’s ability to understand and manipulate tabular data.

To address \textit{Challenge 2}, \textbf{2D LoRA} is designed to address the limited information derived from the two-dimensional cell positions compared to the rich semantics each token conveys. To tackle this, we encode row and column indices using low-rank embeddings and upscale these to integrate with the Large Language Model's (LLM) token embeddings. This approach provides precise row and column index identifiers, enabling the LLM to infer whether two cells align along the same row or column. The importance of this structural awareness cannot be overstated for tasks that require the comprehension of tabular data. The 2D LoRA operates in parallel with the original LoRA framework for each layer, enhancing the LLM's ability to effectively incorporate structural information and to generate content based on structured tabular data.

We conducted experiments on three models across four datasets that encompass QA and fact verification tasks on the tables. The results indicate that TableLoRA consistently demonstrates improvements over vanilla LoRA. Specifically, it achieves a 5.9\% improvement in HiTab. Furthermore, TableLoRA mitigates 40.56\% of the performance gap between LoRA and full fine-tuning. 
For further validation, we conducted control experiments contrasting various table representation learning methods, highlighting the advantages of TableLoRA's structural design. Finally, we designed and executed a series of exploratory analytical experiments to unravel the principles underlying TableLoRA's efficacy.

The main contributions are as follows:

\begin{itemize}[leftmargin=0pt,itemindent=\parindent, itemsep=2pt,topsep=0pt,parsep=0pt]
\item TableLoRA is the first to propose a table-specific LoRA, which aids in learning structured information from tables by modifying the model architecture. This innovative approach enhances the model's ability to understand and process tabular data effectively.

\item  Under PEFT low-parameter settings, TableLoRA demonstrates a superior capability to capture and learn table structures. This makes it highly efficient and effective in scenarios where computational resources and parameter budgets are limited.

\item  The experimental results of TableLoRA are impressive. It consistently outperforms vanilla LoRA, showing significant improvements in various datasets, thus validating its efficacy and robustness in handling table-related tasks.

\item  We summarized and designed various methods for table representation learning, and, through controlled experiments, we proved that the current TableLoRA is the optimal solution among them.
\end{itemize}

\begin{figure*}[htbp]
    \centering
    \includegraphics[width=0.9\linewidth]{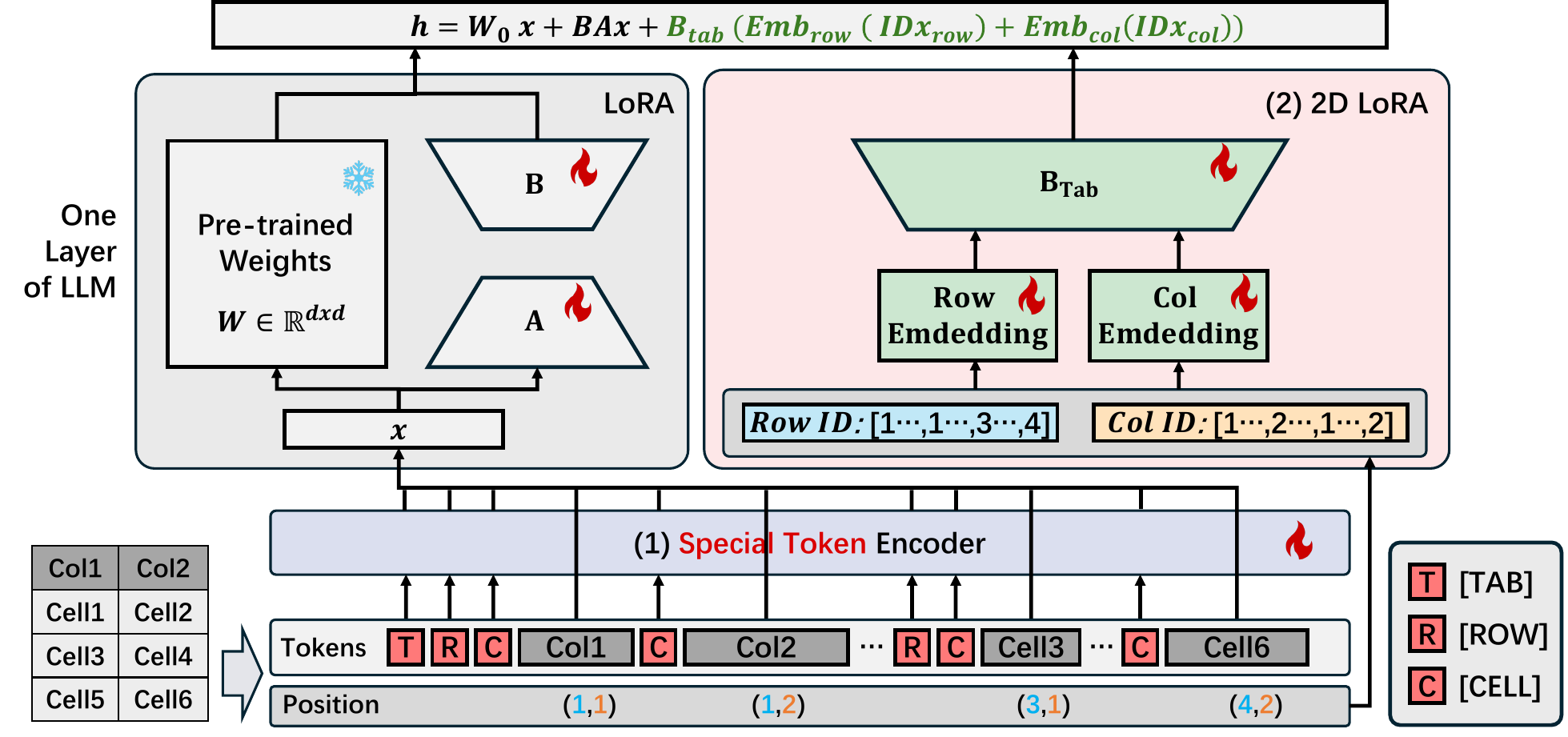}
    \caption{The Framework of TableLoRA. It consists of two main components: Special Tokens Encoder and 2D LoRA. (1) The Special Tokens Encoder (\refsec{sec:encoder}) incorporates specially defined tokens into the model's input alongside word embeddings just before the transformer layers. (2) The 2D LoRA (\refsec{sec:2d_lora}) embeds row and column indices and integrates them into the model at each layer, enabling the processing of tabular data's structure and content.}
    \label{fig:table_lora}
    \vspace{-3mm}
\end{figure*}

\section{Related Work}
\subsection{Tabular Task}
Recently, table-related tasks have received considerable attention in the field. Various tasks and datasets have been proposed, such as TableQA (WikiTQ~\cite{wikitq}, FeTaQA~\cite{Nan2021FeTaQAFT}, HiTab~\cite{cheng-etal-2022-hitab}, HybridQA~\cite{chen-etal-2020-hybridqa}), Text2SQL (WikiSQL~\cite{zhongWikiSQL2017}, Spider~\cite{spider}, BIRD~\cite{bird}), Fact Verification (TabFact~\cite{2019TabFactA}), and tabular analysis (Table2Analysis~\cite{Zhou2020Table2AnalysisMA}, Table2Charts~\cite{zhou2021charts}, AnaMeta~\cite{he-etal-2023-anameta}, DS-1000~\cite{Lai2022DS1000}, Text2Analysis~\cite{text2analysis}). These datasets include various types of tables, such as database tables (\eg, BIRD~\cite{bird}, Spider~\cite{spider}), simple tables (tables with the first row as a header and several subsequent rows containing corresponding values, \eg, WikiTQ~\cite{wikitq}), and hierarchical tables (tables with tree-structured hierarchy in the top header or the left header, \eg, HiTab~\cite{cheng-etal-2022-hitab}). Their common feature is that the input includes tables, requiring models to have the capability to understand and analyze the tables, such as understanding row-column correspondences and positional information within the tables.

\subsection{Tabular Representation Learning}
\label{sec:relate_tab_method}
In the era preceding Large Language Models, table representation methodologies can be divided into model designs and extra training techniques~\cite{ijcai2022tablepretraining}. Within model designs, various approaches are employed to effectively capture the structural details of tabular data:
(1) Table Serialization: This method linearizes tables for better integration with transformer-based models, \eg, TaPEx~\cite{liu2021tapex}, TABBIE~\cite{iida2021tabbie}.
(2) Structural Positional Encoding: This technique encodes structural information, such as row and column indices, to preserve spatial relationships within tables \eg, TaPas~\cite{herzig2020tapas}, MATE~\cite{eisenschlos2021mate}, TABBIE~\cite{iida2021tabbie}, and TUTA~\cite{wang2021tuta}.
(3) Structure-Based Attention Mechanisms: These mechanisms incorporate structural information into the model through attention, enhancing the focus on relevant table components, \eg, TURL~\cite{deng2020turl} and TUTA~\cite{wang2021tuta}.
(4) Multiple Encoder Frameworks: This approach uses multiple encoders to process table data more comprehensively, \eg, TABBIE~\cite{iida2021tabbie}, DoT~\cite{krichene2021dot}, and KGPT~\cite{chen2020kgpt}.
For the design of our control experiments in \refsec{sec:control}, we draw inspiration from these model design strategies. However, it has been observed that directly applying these methods to the era of LLMs does not bring significant improvements. 

\subsection{Tabular Large Language Model}
With the advent of the era of LLMs, a series of works related to large models for tables have followed one after another. To enhance the performance of LLMs in table tasks, existing work primarily focuses on improvements from the following perspectives: (1) Input sequence level: employing different serialization or augmentation methods, \eg, TAP4LLM~\cite{sui-etal-2024-tap4llm}, SpreadsheetLLM~\cite{dong2024spreadsheetllm}, CoCoST~\cite{he-etal-2024-cocost}; (2) Reasoning level: increasing task accuracy by breaking down problems into multiple steps to form a reasoning chain, \eg, Chain-of-Table~\cite{wang2024chain}; (3) Training strategy level: enhancing training effects by constructing large-scale training datasets and synthesizing data, \eg, TableLlama~\cite{zhang2023tablellama}, TableLLM~\cite{zhang2024tablellm}, StructLM~\cite{zhuang2024structlm}, Table-GPT~\cite{li2024table-gpt}. However, in scenarios with limited data and computational resources, there is a lack of relevant exploration on how to more efficiently improve the model's capabilities in table tasks. Moreover, in the PEFT setting, merely modifying the input sequence does not sufficiently enable the model to learn the structural information of tables. Therefore, in TableLoRA, we have made improvements specifically for tables through model design.

\subsection{Parameter-Efficient Fine-Tuning}
Parameter-Efficient Fine-Tuning (PEFT) methods efficiently adapt large pretrained models to various downstream tasks by fine-tuning only a small subset of additional parameters, significantly reducing computational and storage costs while maintaining performance. PEFT encompasses techniques like soft prompt methods (\eg, Prompt Tuning~\cite{lester-etal-2021-prompttuning}, P-tuning~\cite{p-tuning}), which optimize specific task parameters by adding learnable prompts to the input embeddings. Another approach is low-rank adaptation (\eg, LoRA~\cite{hu2022lora}), which inserts smaller trainable matrices into the model. Despite their advantages, these methods lack the ability to understand table structures. TableLoRA addresses this and improve performance on tasks involving structured data.

\section{Methodology}

Tabular tasks involve generating an answer sequence $output$ given a table $T$ and related text $text$ (such as questions, table captions, etc.). The table consists of $n$ columns and $m$ rows of cells, $T=\{v_{0,0}, v_{0,1}, \ldots, v_{m,n}\}$.

TableLoRA is divided into two primary components as shown in \reffig{fig:table_lora}: Special Tokens Encoder and 2D Low-Rank Adaptation (2D LoRA).  The Special Tokens Encoder is designed to enhance the model's understanding of tabular data by incorporating specially defined tokens that provide a clear and structured representation of tables. These special token encoder is added to the model's input at the same stage as the word embeddings, just before the transformer layers. On the other hand, 2D LoRA is aimed at encoding the structural information of tables by embedding row and column indices and integrating them into the Large Language Model at each layer. This dual approach ensures that the model can effectively process both the content and the structure of tabular data, leading to improved performance in related tasks.

\subsection{Special Tokens Encoder}  
  
In special tokens encoder, we introduce the incorporation of special tokens to enhance the representation of tabular data during the fine-tuning process.

We define three special tokens: $\texttt{[tab]}$, $\texttt{[row]}$, and $\texttt{[cell]}$, which are used to replace traditional markdown or HTML formatting. Before entering table $T$ and text $text$ into the model, the table needs to be serialized and concatenated with the text to form a single input sequence $input$. During the serialization process, these special tokens are used as delimiters: $\texttt{[tab]}$ signifies the beginning of a table, $\texttt{[row]}$ indicates the start of a new row, and $\texttt{[cell]}$ marks the beginning of a new cell within a row, as shown in \refequ{equ:serialize}. These tokens are designed to provide a structured and clear representation of tabular data, facilitating more effective processing and comprehension by the model.

{
\small
\begin{equation}
\left\{
    \label{equ:serialize}
    \begin{array}{l}
        input = concate(table\_string, text)\\
        table\_string \\
        \quad = serialize(T)\\
        \quad = \texttt{[TAB]} \ \texttt{[ROW]}\ \texttt{[CELL]}\ v_{0,0}\ \texttt{[CELL]}\ v_{0,1} \ \ldots \ \\
        \quad \quad \texttt{[ROW]}\ \texttt{[CELL]}\ v_{1,0}\ \ldots \ \texttt{[CELL]}\ v_{m,n}
    \end{array}
\right.
\end{equation}
}

To ensure effective learning of special token embeddings, we designed a special token encoder inspired by soft prompt methods~\cite{soft_prompt}, which optimize task-specific parameters by adding learnable prompts to input embeddings. Both approaches require adding new learnable embeddings to the pre-trained model, but the key difference is that soft prompt methods add prompts only at the sequence's beginning, while in TableLoRA, special tokens appear at multiple positions throughout the sequence. Thus, we adopt the encoder from the soft prompt method and extend it to create a position-flexible special token encoder. The formula is as follows:

{
\small
\begin{equation}
\begin{array}{l}
    word\_embedding_i \\
    \quad\quad = 
    \left\{
    \begin{array}{ll}
    WordEmbedding(t_i),  &  t_i \notin S\\
    SpecialTokenEncoder(t_i),     & t_i \in S
    \end{array}
    \right. 
\end{array}
\end{equation}
}
where, $t_i \in input$ is one token of input, $S = \{\texttt{[TAB]}, \texttt{[ROW]}, \texttt{[CELL]}\}$ is the set of special token. Each $word\_embedding_i$ is concatenated to form a word embedding sequence that serves as input to the model.

In this paper, the encoder from P-tuning~\cite{p-tuning} is chosen as the special token encoder, which consists of a word embedding layer and a linear layer. Furthermore, we compare the effects of different soft prompt methods in controlled experiments (\refsec{sec:exp_encoder}).
\label{sec:encoder}
  
\subsection{2D Low-Rank Adaptation (2D LoRA)}  
\label{sec:2d_lora}
Compared to the rich semantics conveyed by each token, the information that can be derived from the 2D cell positions is relatively limited. To address this, we add index embeddings for the row and column indices. Since the information density of these indices is relatively low, we choose to represent them using low-rank embeddings. These low-rank embeddings are then upscaled and integrated with the token embeddings of the LLM. This approach, termed 2D LoRA, operates in parallel with the original LoRA~\cite{hu2022lora} framework for each layer, enabling the LLM to incorporate structural information effectively. Specifically, precise row and column index identifiers are provided, allowing the LLM to infer whether two cells are aligned along the same row or column. This structural awareness is critical for tasks that require an understanding of tabular data.

The mathematical formulation of this integration can be expressed as follows:  
{\small
\begin{equation}
\begin{array}{ccl}
h& = & W_0 x + B A x \\
&&+ B_{\text{tab}} \left( \text{Emb}_{\text{row}}(\text{IDx}_{\text{row}}) \right. + \left. \text{Emb}_{\text{col}}(\text{IDx}_{\text{col}}) \right)
\end{array}
\label{equ:2d_lora}
\end{equation}  
}
where, \(h\) represents the hidden states. \(W_0\), \(B\) and \( A \) are the weight matrix of the LLM and the original LoRA. \(B_{\text{tab}}\) is additional parameters introduced by the 2D LoRA. \(\text{Emb}_{\text{row}}\) and \(\text{Emb}_{\text{col}}\) denote the embeddings of the row and column indices, respectively. \(\text{IDx}\) represents which row/column this token belongs to. Specifically, for tokens that are not part of the table, both the $\text{IDx}_{\text{col}}$ and $\text{IDx}_{\text{row}}$ are set to 0. For special tokens, the token $\texttt{[TAB]}$ has both $\text{IDx}_{\text{col}}$ and $\text{IDx}_{\text{row}}$ set to 0, the token $\texttt{[ROW]}$ has $\text{IDx}_{\text{col}}$ set to 0 and $\text{IDx}_{\text{row}}$ set to the corresponding row number, and the token $\texttt{[CELL]}$ has both $\text{IDx}_{\text{col}}$ and $\text{IDx}_{\text{row}}$ corresponding to its respective cell value.

The origin of 2D LoRA lies in the addition of column ID and row ID embeddings to word embeddings, similar to Tapas~\cite{herzig2020tapas}. The uniqueness of 2D LoRA lies in its specific embeddings, which are distinct for each transformer layer. This differentiation is crucial as layers vary in depth and informational needs, as will be demonstrated in subsequent experiments (see \reffig{fig:deep}), showing varying impacts across layers. Additionally, our use of low-dimensional embeddings to represent row/column IDs captures essential positional information efficiently, without requiring high-dimensional semantic embeddings.

\begin{table*}[htbp]
  \centering
  \small
  \caption{Main Experiments Results. All metric numbers are in \%. $\Delta$ represents the difference between the respective metric and TableLoRA, with red indicating negative values, showing how much it decreased.}
    \resizebox{0.8\textwidth}{!}{
    \begin{tabular}{cl|cc|cc|cc|cc}
    \toprule
    \multicolumn{2}{c|}{\multirow{2}[4]{*}{Model}} & \multicolumn{2}{c|}{HiTab} & \multicolumn{2}{c|}{WikiTQ} & \multicolumn{2}{c|}{FeTaQA} & \multicolumn{2}{c}{TabFact} \\
\cmidrule{3-10}    \multicolumn{2}{c|}{} & acc   & $\Delta$     & acc   & $\Delta$     & bleu  & $\Delta$     & acc   & $\Delta$ \\
    \midrule
    \multirow{3}[2]{*}{Llama 2} & Full Finetune & 48.61  & \textcolor[rgb]{ 1,  0,  0}{-0.33 } & 46.20  & 5.74  & 32.10  & 4.10  & 52.29  & \textcolor[rgb]{ 1,  0,  0}{-25.77} \\
          & Lora  & 43.00  & \textcolor[rgb]{ 1,  0,  0}{-5.94 } & 38.76  & \textcolor[rgb]{ 1,  0,  0}{-1.70 } & 25.13  & \textcolor[rgb]{ 1,  0,  0}{-2.87 } &   76.93    & \textcolor[rgb]{ 1,  0,  0}{-1.13} \\
          & TableLoRA & 48.94  & 0.00  & 40.46  & 0.00  & 28.00  & 0.00  &   78.05    & 0.00 \\
    \midrule
    \multirow{3}[2]{*}{Llama 3} & Full Finetune & 61.62  & 3.05  & 56.84  & 3.39  & 34.86  & 4.62  & 84.18  & 0.17  \\
          & Lora  & 57.06  & \textcolor[rgb]{ 1,  0,  0}{-1.50 } & 51.98  & \textcolor[rgb]{ 1,  0,  0}{-1.47 } & 29.09  & \textcolor[rgb]{ 1,  0,  0}{-1.14 } & 83.49  & \textcolor[rgb]{ 1,  0,  0}{-0.52 } \\
          & TableLoRA & 58.56  & 0.00  & 53.45  & 0.00  & 30.23  & 0.00  & 84.01  & 0.00  \\
    \midrule
    \multirow{3}[2]{*}{DeepSeek} & Full Finetune & 54.92  & 7.99  & 47.01  & 6.59  & 32.15  & 4.87  & 79.44  & 2.40  \\
          & Lora  & 43.25  & \textcolor[rgb]{ 1,  0,  0}{-3.69 } & 37.34  & \textcolor[rgb]{ 1,  0,  0}{-3.08 } & 26.68  & \textcolor[rgb]{ 1,  0,  0}{-0.61 } & 75.20  & \textcolor[rgb]{ 1,  0,  0}{-1.85 } \\
          & TableLoRA & 46.94  & 0.00  & 40.42  & 0.00  & 27.29  & 0.00  & 77.05  & 0.00  \\
    \midrule
    \multirow{3}[2]{*}{Large Models} & GPT-4o & 55.05 & - & 58.38 & - & 14.34 & - & 44.01 & - \\
& Claude-3.7 & 56.88 & - & 66.90 & - & 13.58 & - & 80.14 & - \\
& Llama 3.3-70B & 22.79 & - & 41.30 & - & 7.66 & - & 15.99 & - \\
    \bottomrule
    \end{tabular}%
    }
  \label{tab:main}%
  \vspace{-3mm}
\end{table*}%

\section{Experiment}

% We conducted experiments on four tabular-related datasets (HiTab, WikiTQ, FetaQA, TabFact) that cover table QA and fact verification tasks. Furthermore, we experimented with three models, Llama2, Llama3, and DeepSeek. 
We conducted three parts of the experiment: First, the main experiment, which involved performing TableLoRA and baseline experiments on 3 models in 4 datasets to demonstrate the effectiveness of TableLoRA. Second, the control experiment, in which we designed various variants of encode table methods to compare with TableLoRA, highlighting the advantages of TableLoRA's structural design. Third, further analysis, including an ablation study and in-depth analysis of TableLoRA, to deeply explore the mechanisms of TableLoRA.

\subsection{Experiment Setup}

% \subsubsection{Datasets}

Our experiments were conducted on four table-related datasets: HiTab~\cite{cheng-etal-2022-hitab}, WikiTQ~\cite{wikitq}, FeTaQA~\cite{Nan2021FeTaQAFT}, and TabFact~\cite{2019TabFactA}. The first three are Table QA datasets, where the input consists of a table and a related query, and the task is to answer the query based on the table, with the output being the answer to the question. Among them, HiTab involves tables with complex hierarchical structures, such as multi-layered tree structures in the top or left header, as shown in \reffig{fig:error_eg}. The last dataset, TabFact, is for fact verification, where the input is a table and a related statement, and the task is to determine the truthfulness of the statement based on the table, with the output being the judgment result.
% \TODO{statistics}

% \subsubsection{Base LLMs}

We conducted experiments on three open-source LLMs: Llama 2~\cite{touvron2023llama2openfoundation}, Llama 3, and DeepSeek~\cite{deepseek-llm}. In addition, we include several larger models as baselines for comparison. Due to their closed-source nature or computational constraints, we perform inference only on these models without fine-tuning. Details are provided in \refsec{sec:app_model}.
We use the official metrics for each dataset, as shown in the header of \reftab{tab:main}. 
% HiTab, WikiTQ, and TabFact use accuracy, and FeTaQA uses BLEU.

All experiments were run on Linux machines with 4 NVIDIA Tesla A100 80G memory GPUs. The LLaMA Factory framework served as the foundation, which we extensively customized to incorporate TableLoRA-related techniques and methods. In the main experiments, LoRA and 2D LoRA share the same rank (8). PEFT is applied to the $k\_proj$ and $v\_proj$ layers. For more training details, see \refsec{sec:app_exp}.

\subsection{Main Results}
\label{sec:main_exp}
In the main experiments, we evaluated TableLoRA, the original LoRA~\cite{hu2022lora}, and full parameter fine-tuning across three models and four datasets. Results are presented in \reftab{tab:main}, and experimental parameters are detailed in \refsec{sec:app_exp}. Few have attempted to improve PEFT for table tasks, so no additional baselines are available. To validate the model structure's superiority, we summarized common table representation methods and designed variant experiments for comparison in \refsec{sec:control}.

The results show that TableLoRA outperforms baseline LoRA fine-tuning, with a 5.9\% improvement on the Llama2 model when applied to the HiTab dataset, demonstrating its effectiveness with table-based inputs.

In a low-parameter setting, TableLoRA mitigates LoRA’s shortcomings compared to full fine-tuning for table tasks. It reduces the performance gap between LoRA and full fine-tuning by an average of 40.56\% (excluding the TabFact outlier on Llama2, as detailed in \refsec{sec:app_anomaly}). Specifically, on the HiTab dataset with Llama2, TableLoRA improves LoRA by 5.95\%, matching full fine-tuning performance. This demonstrates that TableLoRA can learn complex tabular structures with fewer parameters, addressing LoRA’s deficiencies.

Compared with large models, on WikiTQ, Claude 3.7 and GPT-4o generally outperform TableLoRA and related baselines, indicating that tuning-based approaches still require improvement compared to reasoning-focused LLMs, and may need enhancement through incorporation of reinforcement learning for further progress. In contrast, on FeTaQA, TableLoRA and its baselines consistently surpass reasoning LLMs, potentially because the tuning process enables better acquisition of FeTaQA's answer sentence structures, thereby achieving higher BLEU scores.

\subsection{Control Experiments}

\subsubsection{Table Representation learning Methods}
\label{sec:control}

To further validate the superiority of the TableLoRA structure, we summarized common table representation learning methods in \refsec{sec:relate_tab_method} and designed controlled experiments based on those designs. The results are shown in \reftab{tab:control}. Referring to the table pre-training methods before the era of LLMs~\cite{ijcai2022tablepretraining}, common table representation learning methods that can be applied to LLMs include: describing through strings, enhancing structural information through positional embedding, and enhancing structural information through attention masks. Therefore, we designed the following variant experiments, and more details are shown in \refsec{sec:app_control}.

% \TODO{More details}
\textbf{Different format}: Table input can be serialized using various formats, such as markdown, HTML, and CSV (the main experimental baseline in \refsec{sec:main_exp} uses markdown). Compared to the special token encoder in TableLoRA, adding a special token yielded better results than the best-performing markdown format.

\textbf{Add in string sequence}: The positional information in the table (e.g., which row and which column) is added to the string sequence.  Compared with the 2D LoRA in TableLoRA, the latter incorporates positional information into the model through embedding, allowing the model to learn positional information more directly, resulting in greater improvement.

\textbf{Add in positional embedding}: The positional information in the table (e.g., which row and which column) is added to the word embedding through positional embedding. Experiment results show that this does not bring about an effective improvement to the model. Compared with the 2D LoRA in TableLoRA, the latter incorporates positional information into each layer of the model through LoRA, continually emphasizing structural information at different layers during model inference, resulting in greater improvement.

\textbf{Add in attention mask}: This approach improves the model's attention mechanism by incorporating positional information from tables, focusing on two variants: (1) highlighting tokens within the same cell, and (2) highlighting tokens within the same row, column, or cell. In comparison to the 2D LoRA method in TableLoRA, the latter variant learns position vectors in different feature spaces at different layers through embedding, allowing for better integration with the model and resulting in a greater improvement in performance.

% Table generated by Excel2LaTeX from sheet 'Sheet2'
\begin{table}[tb]
  \centering
  \small
  \caption{Control Experiments on the LLama2 of the Hitab. All metric numbers are in \%.}
    \resizebox{0.8\columnwidth}{!}{
    % Table generated by Excel2LaTeX from sheet 'Sheet2'
    \begin{tabular}{lc}
    \toprule
    Control Exp & HiTab \\
    \midrule
    TableLoRA& \textbf{48.94 } \\
    LoRA (same rank) & 43.00  \\
    LoRA (comparable params) & 43.06  \\
    \midrule
    \multicolumn{2}{c}{Special Token Control Exp} \\
    Different format (html) & 32.25  \\
    Different format (csv) & 43.06  \\
    Prompt Tuning & 42.56  \\
    \midrule
    \multicolumn{2}{c}{2D LoRA Control Exp} \\
    Add in string sequence & 31.06  \\
    Add in attention mask (1) & 40.50  \\
    Add in attention mask (2) & 41.00  \\
    Add in positional embedding & 39.00  \\
    \bottomrule
    \end{tabular}%
    }
    \vspace{-3mm}
  \label{tab:control}%
\end{table}%

\subsubsection{Special Token Encoder}
\label{sec:exp_encoder}
We evaluated the impact of different encoders for special tokens. The special token encoder in TableLoRA is inspired by soft prompt methods, as described in \refsec{sec:encoder}. Common soft prompt methods include P-tuning and prompt tuning~\cite{lester-etal-2021-prompttuning}. For the control experiment, we use the \textbf{prompt tuning} encoder, which mainly consists of a word embedding layer. We modify it similarly to P-tuning for our experiments. The results in \reftab{tab:control} show that the prompt tuning encoder fails to effectively learn table-related token embeddings at different positions in the sequence.

\subsubsection{The Number of Parameters}

To verify that the effectiveness of TableLoRA is not due to an increase in the number of parameters, we conducted experiments by increasing the parameter count of the original LoRA to make it comparable with TableLoRA. In Llama2, TableLoRA trains 0.130\% of the parameters, while \textbf{LoRA (same rank)} (the baseline in \refsec{sec:main_exp}) trains 0.062\% of the parameters. By increasing the rank of LoRA from 8 to 16, we obtain \textbf{LoRA (comparable params)}, which trains 0.124\% of the parameters. The experimental results are shown in \reftab{tab:control}.

The experimental results indicate that the performance improvement of TableLoRA is due to its effective encoding of the table structure, rather than to a small increase in the number of parameters. Even when LoRA's parameter count is increased to a comparable level, it only provides a 0.06\% improvement over the same-rank LoRA. In contrast, TableLoRA delivers a 5.88\% improvement over the parameter-comparable LoRA.

\subsection{Further Analysis}
% Table generated by Excel2LaTeX from sheet 'Sheet2'
% \begin{table}[tb]
%   \centering
%   \caption{Ablation Study on the LLama2 of the Hitab. All metric numbers are in \%.}
%   \resizebox{0.9\columnwidth}{!}{
%     \begin{tabular}{cc|c}
%     \toprule
%     Special Token Encoder & 2D LoRA & HiTab \\
%     \midrule
%     $\times$ & $\times$ & 43.00 \\
%     $\times$ & \checkmark  & 47.19 \\
%     \checkmark  & $\times$ & 44.13 \\
%     \checkmark  & \checkmark  & \textbf{48.94} \\
%     \bottomrule
%     \end{tabular}%
%     }
%   \label{tab:ablation}%
% \end{table}%

% Table generated by Excel2LaTeX from sheet 'Sheet2'
% Table generated by Excel2LaTeX from sheet 'Sheet2'
\begin{table}[tb]
\small
  \centering
  \caption{Ablation Study on the LLama2 of the Hitab. All metric numbers are in \%.}
  \resizebox{0.9\columnwidth}{!}{
    \begin{tabular}{l|cc}
    \toprule
    \multicolumn{1}{c|}{Method}  & HiTab & $\Delta$ \\
    \midrule
    TableLoRA & \textbf{48.94} & 0 \\
    \ \ w/o Special Token Encoder & 47.19 & -1.75 \\
    \ \ w/o 2D LoRA  & 44.13 & -4.81 \\
    \ \ LoRA & 43    & -5.94 \\
    \bottomrule
    \end{tabular}%
    }
    \vspace{-3mm}
  \label{tab:ablation}%
\end{table}%

\subsubsection{Ablation Study}
An ablation study is conducted to further investigate the components that contribute to the enhancements in the performance of the model in \reftab{tab:ablation}. The results indicate that both the Special Token Encoder and 2D LoRA can bring about improvements to varying degrees, with 2D LoRA in particular contributing to a 4.81\% enhancement. Moreover, the improvements are more pronounced when both are used simultaneously.

\subsubsection{Tables of Varying Complexity}
\begin{figure}[tb]
    \centering
    \includegraphics[width=\linewidth,trim={2.5cm 0cm 1cm 0cm},clip]{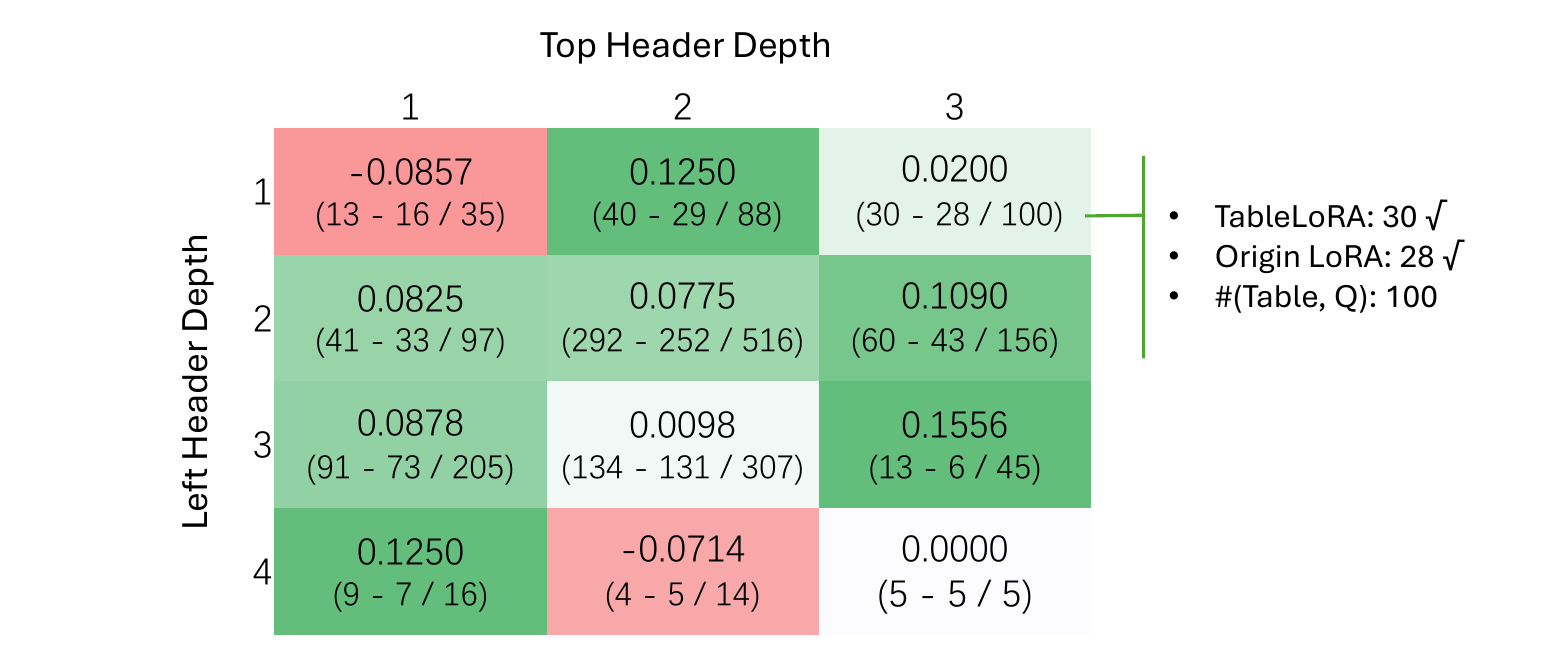}
    \caption{Heatmap of TableLoRA's Enhancement on HiTab Dataset Influenced by Top / Left Header Depth.}
    \label{fig:deep}
    \vspace{-3mm}
\end{figure}

To deeply explore the improvements brought by TableLoRA to tables of different complexities, we tested the enhancements of TableLoRA with varying upper/left header depths on the HiTab dataset~\cite{cheng-etal-2022-hitab}. The header depth refers to the hierarchical depth of headers in a hierarchical table, where 1 represents a single-layer header with no hierarchical structure, and 2 or above indicates the corresponding deepest hierarchical depth. The results are shown in \reffig{fig:deep}, where the values indicate the performance of TableLoRA compared to LoRA.

It can be seen that TableLoRA provides greater improvements on more complex tables with deeper hierarchies (i.e., further to the right or lower in the heatmap). This is because TableLoRA enhances the model's understanding of the table structure, especially the relationships between rows and columns and the hierarchical relationships. However, for particularly complex tables, such as those with top/left header depths of 3/4, the improvements brought by TableLoRA are limited, indicating areas needing further improvement.

When both the top and left header depths are 1, a slight drop is observed. (1) The HiTab dataset is imbalanced, containing only 35 samples with one-hierarchy headers. This small sample size limits the reliability of statistical conclusions. Case studies of the three misclassified examples indicate that the errors were caused by perturbations. (2) To further assess the effectiveness of TableLoRA on flat tables—including those with one-hierarchy headers—we conducted experiments on additional flat-structured datasets (\eg, WikiTQ, FeTaQA, TabFact, see \reftab{tab:main}), where TableLoRA consistently achieves stable improvements.

\subsubsection{Queries with Different Aggregations}
\begin{figure}
    \centering
    \includegraphics[width=\linewidth]{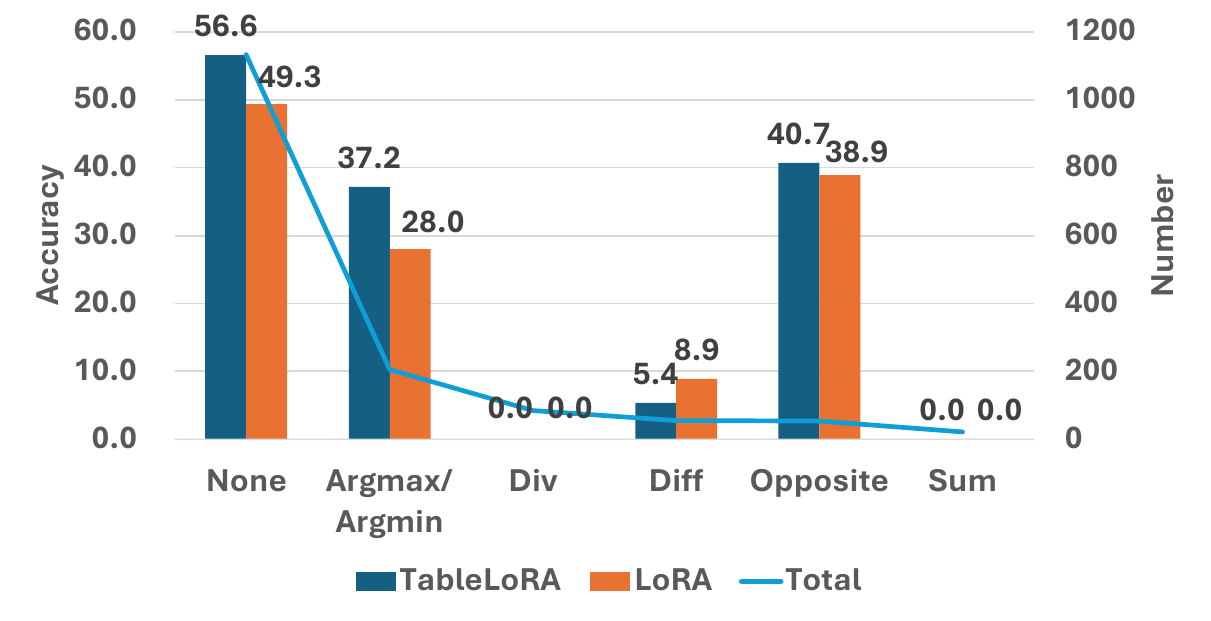}
    \caption{Performance and Data Volumes of Queries with Different Aggregations on HitTab. Accuracy is in \%.}
    \label{fig:aggregation}
    \vspace{-3mm}
\end{figure}

In order to further explore the underlying mechanisms and scope of TableLoRA, we conducted experiments to evaluate the model's performance on various aggregations (functions in HiTab~\cite{cheng-etal-2022-hitab}) involved in the queries, as illustrated in \reffig{fig:aggregation}. To ensure the experiment's reliability, we selected and analyzed aggregation types from the HiTab dataset that had more than 20 samples.

For queries that require precise positioning using table structures, the improvement is more pronounced, such as with Argmax/Argmin and None. TableLoRA showed a 17.2\% improvement on Argmax/Argmin, as it needs to identify which elements in the table should be included in the computation. These elements are usually located in the same row or column, and the 2D LoRA in TableLoRA helps the model to position them more accurately. TableLoRA demonstrated a 7.3\% improvement on ``None'' queries, which do not involve aggregation and are typically for retrieval. TableLoRA enhances the model's ability to locate cell positions and retrieve the results.

The model's ability to perform numerical computations still requires improvement. Regardless of whether TableLoRA is applied, the model performs poorly on queries involving arithmetic operations such as Div, Diff, and Sum, sometimes even achieving 0\% accuracy. Future research should explore how to enhance the model's computational capabilities when working with tables.

\subsubsection{2D LoRA at Different Layers}
\begin{figure}[tb]
    \centering
    \includegraphics[width=0.95\linewidth,trim={0.5cm 0cm 0cm 0cm},clip]{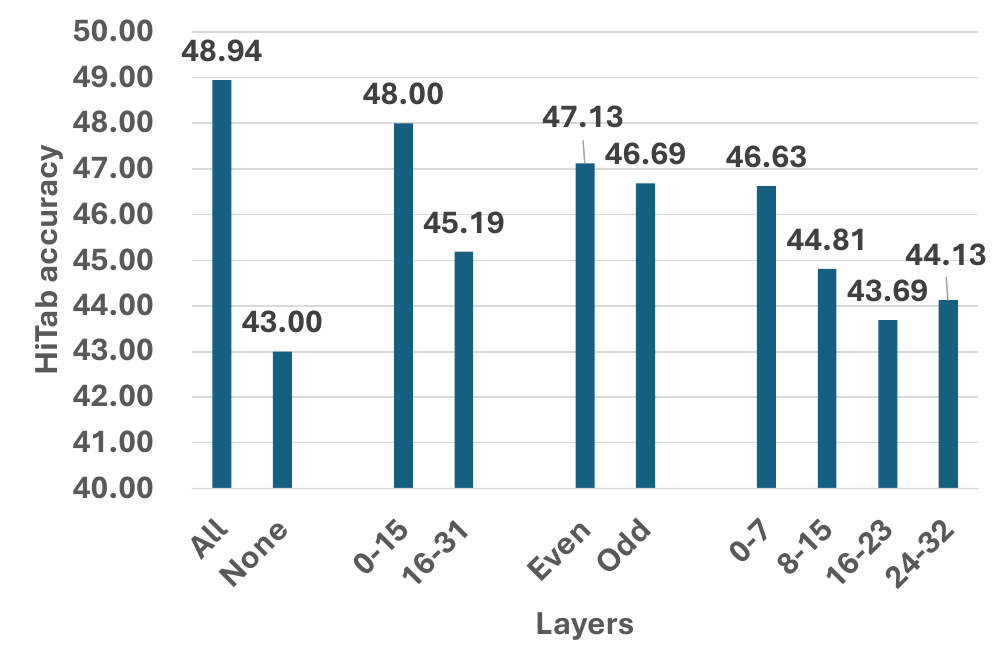}
    \caption{HiTab 2D LoRA Performance Results for Different Layers. All metric numbers are in \%.}
    \label{fig:layers}
    \vspace{-3mm}
\end{figure}

The impact of different layer 2D LoRA on the results is shown in \reffig{fig:layers}. In the experiment, selected model layers use 2D LoRA, while unselected layers use LoRA. Details are in \refsec{sec:app_layer}.

It can be observed that the earlier the layer, the more conducive it is to learning 2D information. For instance, whether the layers are divided into two or four parts, performance decreases as the number of layers increases. Additionally, the even-numbered layers, being one layer ahead of the odd-numbered layers, exhibit better performance compared to the odd-numbered layers.

\subsubsection{TableLoRA with Different Rank}
\begin{figure}[tb]
    \centering
    \includegraphics[width=\linewidth, trim=0 5 0 0, clip]{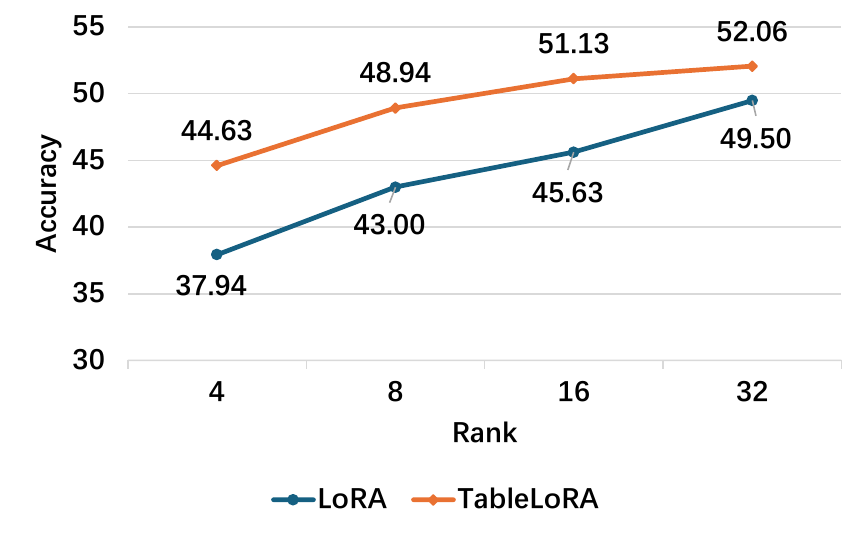}
    \caption{Performance of TableLoRA and LoRA with Different Ranks on HitTab. Accuracy is in \%.}
    \label{fig:rank}
    \vspace{-3mm}
\end{figure}

\reffig{fig:rank} illustrates the performance of TableLoRA and LoRA under different ranks, demonstrating that TableLoRA consistently improves with varying ranks, indicating the scalability of the method.

\subsubsection{Case Study}

The improvements brought by TableLoRA are concretely visualized in \reffig{fig:error_eg}. As shown, the model is enhanced primarily in two aspects. First, the row-column correspondence has been improved. For instance, in the column selection of the query in \reffig{fig:error_eg}, the original query can accurately identify the "canadian-born" column. However, due to LoRA’s error in detecting which cells belong to the same column, the final retrieved result is incorrect. Second, TableLoRA improves the understanding of the tree-structured header. For example, when performing row selection in the query in \reffig{fig:error_eg}, LoRA selects the nearest upper row for "15 to 24 years" without recognizing that it is a parent node in a hierarchical structure, with its child nodes corresponding to several rows below. By addressing these two issues, TableLoRA successfully retrieves the correct answers corresponding to the query.

\section{Conclusion}
In summary, this paper presents TableLoRA, a novel method for enhancing LLMs' understanding of tabular data within the PEFT. By introducing special tokens encoder for table serialization and a 2D LoRA mechanism to encode cell positions, TableLoRA addresses the structural comprehension limitations of existing models. Experiments on multiple datasets show that TableLoRA consistently outperforms vanilla LoRA and the other table representation learning methods, demonstrating significant improvements in handling table-related tasks. This approach is both efficient and effective, marking a significant advancement in the fine-tuning of LLMs for tabular data.

\section*{Limitations}
One limitation of this paper is that we can only validate our method on open-source models. Unfortunately, we are unable to test its effectiveness on the current state-of-the-art GPT series models due to accessibility constraints. This limitation reduces the generalizability of our findings to the most advanced models available. Additionally, the experiments require substantial GPU computational resources, raising concerns about energy consumption and environmental sustainability.

Another limitation is that, although TableLoRA demonstrates stable improvements compared to LoRA, it still cannot match the results achieved through full finetuning. This discrepancy suggests that while TableLoRA offers notable benefits, it may not yet reach the performance level achieved by fine-tuning techniques.

\section*{Ethics Statement}
The datasets and other associated resources utilized in this study are publicly available and widely used in various other existing work.
All the datasets used in this paper have been reviewed to ensure that they do not contain personally identifiable information or offensive content. 
However, since these datasets are sourced from the Internet, potential bias may still be present. 
Furthermore, despite our careful review, the process involving the LLMs may inadvertently introduce inappropriate information into the evolved data. 

\section*{Acknowledgments}
We thank all anonymous reviewers for their valuable comments. Xinyi He and Zejian Yuan were supported in part by the National Key R\&D Program of China (2023YFB4704900) and NSFC (61976170, 62088102).

\bibliography{custom}

\clearpage
\appendix

\section{Experiment Configuration Details}
\subsection{Datasets Selection}
We selected four datasets (HiTab, WikiTQ, FeTaQA, TabFact) because they are classic, reliable, and cover diverse table structures and tasks. \reftab{tab:dataset} summarizes their characteristics. Varying instruction systematically test the model’s ability to locate and retrieve information from tables at different difficulty levels, while diverse output rigorously assess its capacity to generate responses with varying linguistic or structural demands.

\begin{table}[htbp]
\small
  \centering
  \caption{Dataset Selection.}
  \resizebox{0.9\columnwidth}{!}{
\begin{tabular}{c|ccc}
\hline
Dataset & Table        & Instruct       & Output    \\ \hline
HiTab   & hierarchical & query          & words     \\
WikiTQ  & flat         & query          & words     \\
FeTaQA  & flat         & query          & sentences \\
TabFact & flat         & fact/statement & bool      \\ \hline
\end{tabular}
    }
  \label{tab:dataset}%
\end{table}%

\subsection{Models Used}
\label{sec:app_model}
For all experiments conducted in this paper, we employed three pre-trained large language models (LLMs): DeepSeek LLM-7B Chat (deepseek-ai/deepseek-llm-7b-chat), Llama 2-7B Chat (meta-llama/Llama-2-7b-chat-hf), and Meta Llama 3-8B Instruct (meta-llama/Meta-Llama-3-8B-Instruct). 
Llama 2 is a collection of pre-trained and fine-tuned models optimized for dialogue, outperforming most open-source chat models. Llama 3, the next generation of Llama models, offers enhanced reasoning capabilities and strong performance across various benchmarks. DeepSeek is an advanced model trained on a large English and Chinese dataset, available in both base and chat versions, and open-sourced for the research community. 
These models were selected based on their relevance in the current landscape of language model research and their suitability for fine-tuning tasks on structured datasets, such as the TabFact dataset.

\section{Training Configurations}
\label{sec:app_exp}

\subsection{Hardware and Frameworks}
We employed machines with four NVIDIA A100 GPUs for fine-tuning. The LLaMA Factory framework served as the foundation, which we extensively customized to incorporate TableLoRA-related techniques and methods. To enable full-parameter fine-tuning of large-scale models, we used DeepSpeed version 0.14.4. The configuration file employed during fine-tuning was the default DeepSpeed zero-2 stage configuration file from the example directory provided by the LLaMA Factory framework. To ensure consistency and eliminate any framework-induced bias in the results, we applied the same DeepSpeed framework and configuration file for both LoRA fine-tuning and TableLoRA fine-tuning.

\subsection{LoRA Fine-Tuning}
The LoRA fine-tuning used eight LoRA ranks with an alpha value of 16 and a dropout rate of 0.1. In all cases, training employed a batch size of 8 per device, with gradient accumulation steps of 2, a learning rate of 5e-6, and a cosine scheduler. Training was conducted for three epochs with a maximum sequence length of 4,000 tokens on the TabFact dataset and 1,000 tokens on other datasets. Mixed precision (FP16) and distributed training were enabled using DeepSpeed. TableLoRA, a variant of LoRA, used the same fine-tuning parameters.

\subsection{TableLoRA Fine-Tuning}
The main implementation of TableLoRA involves both Special Tokens and 2D LoRA. The training of Special Tokens employs p-tuning with its default parameters, while the hyperparameters for 2D LoRA are consistent with those of LoRA. Specifically, for 2D LoRA, the maximum values for columns and rows are set to 40 and 600 on other datasets, and 50 and 600 on the TabFact dataset.

\subsection{Full-parameter Tuning}
The full fine-tuning process employed a learning rate of 5e-6 with a cosine scheduler and a maximum gradient norm of 1.0. Training was conducted over 3 epochs with a maximum sequence length of 4000 tokens, utilizing a batch size of 8 per device and gradient accumulation steps of 2. The model training leveraged mixed precision (FP16) and distributed training capabilities provided by DeepSpeed. To ensure efficiency and stability, the preprocessing pipeline involved 16 workers, and warmup steps were set to 0. Additionally, the training process included advanced optimization techniques such as AdamW, with careful monitoring of loss curves and model performance metrics throughout.

\subsection{Consistency in Training}
To ensure comparability across different models and datasets, we applied the same training configurations to all experiments. This uniformity minimized the influence of hyperparameter differences, isolating the effects of model architectures and dataset characteristics. By maintaining consistent training parameters, we could confidently attribute variations in performance to the intrinsic properties of the models or datasets rather than external factors.

All models demonstrated expected convergence behaviors, with the training process yielding the lowest observed loss values for each model. This result confirmed the stability and reliability of the training procedures. The standardized and well-calibrated configurations enabled us to conduct a robust comparison across different models and fine-tuning techniques, ultimately producing meaningful and consistent insights.

\section{Details of Main Results}

\label{sec:app_anomaly}
To quantify the average gap reduction between different fine-tuning methods, we calculate the relative improvement of the TableLoRA method over the LoRA method, normalized by the difference between the full finetune method and the LoRA method. Mathematically, this can be expressed as:  

% {\small
\begin{equation*}  
\text{Gap Reduction} = \frac{1}{n} \sum_{i=1}^{n} \frac{P_{\text{full\_finetune}_i} - P_{\text{tablelora}_i}}{P_{\text{full\_finetune}_i} - P_{\text{lora}_i}}  
\end{equation*}  
% }

where \( P_{\text{tablelora}_i} \) represents the performance metric obtained using the TableLoRA method, \( P_{\text{lora}_i} \) represents the performance metric obtained using the LoRA method, and \( P_{\text{full\_finetune}_i} \) represents the performance metric obtained using the full finetune method for the \( i \)-th instance. The total number of instances is denoted by \( n \). 

When fine-tuning LLaMA2 on the TabFact dataset, we observed that full-parameter tuning resulted in significantly lower accuracy compared to LoRA, despite both approaches being applied to the same task. One possible reason for this is that LLaMA2's pretraining may not be well-aligned with the task-specific requirements of TabFact, particularly in terms of logical reasoning and table-based data modeling. Full-parameter fine-tuning, which adjusts all weights, might inadvertently interfere with the model’s pre-existing knowledge, disrupting its ability to generalize effectively. On the other hand, LoRA’s approach, which only adjusts a small set of parameters, focuses more on task-specific patterns, leading to better performance. Furthermore, we encountered issues related to optimization during full-parameter fine-tuning, such as gradient vanishing, which made the optimization process unstable. This instability often led to convergence problems, preventing the model from reaching an optimal solution. LoRA, due to its reduced parameter space, was less prone to such issues and exhibited a more stable convergence. Additionally, TabFact's inherent noise and specific patterns could have been more effectively captured by LoRA, as it is less likely to overfit to irrelevant features. In contrast, other models like LLaMA3 and DeepSeek may have better adapted to the task during pretraining, resulting in higher accuracy when subjected to full-parameter fine-tuning.

\section{Details of Control Experiment}
\textbf{Different format}: When comparing with the special token encoder, three types of table serialization formats are involved: markdown, HTML, and CSV. In the experiment, we maintained consistency with the pandas library for specific serialization methods: \texttt{DataFrame.to\_markdown()}, \texttt{DataFrame.to\_html()}, and \texttt{DataFrame.to\_csv()}.  

\textbf{Add in string sequence}: We add the position information of each cell to the string sequence. Specifically, the position string ``(row\_idx, col\_idx)'' of each row and column is added to the front of the string of each serialized table cell.

\textbf{Add in positional embedding}: We use Sinusoidal Positional Embedding to encode the row/column indices separately. The row and column indices are consistent with those in \refequ{equ:2d_lora}. The calculated embeddings are added to the word embeddings, similar to the original positional embeddings in the transformer, for computation.

\begin{table*}[tb]
  \centering
  \caption{Ablation Study on each model for each dataset. All metric numbers are in \%.}
  \resizebox{0.75\textwidth}{!}{
    \begin{tabular}{cl|c|c|c|c}
    \toprule
    \multicolumn{2}{c|}{Model} & HiTab & WikiTQ & FeTaQA & TabFact \\
    \midrule
    \multirow{4}[2]{*}{Llama 2} & TableLoRA & \textbf{48.94 } & \textbf{40.46 } & \textbf{28.00 } & \textbf{78.05 } \\
          & w/o Special Token Encoder & 47.19  & 39.73  & 27.72  & 78.00  \\
          & w/o 2D LoRA & 44.13  & 39.60  & 25.91  & 77.37  \\
          & LoRA  & 43.00  & 38.76  & 25.13  & 76.93  \\
    \midrule
    \multirow{4}[2]{*}{Llama 3} & TableLoRA & \textbf{58.56 } & 53.45  & \textbf{30.23 } & \textbf{84.01 } \\
          & w/o Special Token Encoder & 58.43  & \textbf{53.52 } & 29.81  & 83.88  \\
          & w/o 2D LoRA & 57.25  & 52.75  & 30.19  & 83.23  \\
          & LoRA  & 57.06  & 51.98  & 29.09  & 83.49  \\
    \midrule
    \multirow{4}[2]{*}{DeepSeek} & TableLoRA & \textbf{46.94 } & \textbf{40.42 } & 27.29  & \textbf{77.05 } \\
          & w/o Special Token Encoder & 45.63  & 38.87  & \textbf{27.50 } & 76.78  \\
          & w/o 2D LoRA & 44.44  & 37.71  & 26.92  & 75.62  \\
          & LoRA  & 43.25  & 37.34  & 26.68  & 75.20  \\
    \bottomrule
    \end{tabular}%
    }
  \label{tab:ablation_all}%
\end{table*}%

\textbf{Add in attention mask}: The positional attention mask is combined with the causal mask typically used in LLMs, ensuring that causal constraints are preserved while embedding the structural information of the table. This integration occurs at each layer during the forward pass, enabling the model to consistently emphasize table-specific structural patterns throughout inference. At the core of this method is the concept of weight amplification, which boosts the model’s focus on structural information. Tokens within the same cell receive higher attention weight (e.g., adding a mask value of 1), prompting the model to prioritize these tokens. Tokens within the same row or column receive a lower amplification (e.g., mask value of 0.5), highlighting their contextual relevance to a lesser extent, while tokens not sharing a row or column relationship receive no additional weight, maintaining neutral attention scores.
\label{sec:app_control}

\section{Details of Further Analysis}
\subsection{Ablation Study}
The ablation results for each dataset are shown in \reftab{tab:ablation_all}.

\subsection{Performance of 2D LoRA at Different Layers}
\label{sec:app_layer}
We conduct the following sets of experiments for different model layers: (1) Halving: The model layers are divided into two halves for the experiment, i.e., layers 0-15 and 16-31. (2) Odd and Even: The model layers are divided into odd and even numbers for the experiment, i.e., layers 0, 2, ..., 30 and layers 1, 3, ..., 31. (3) Quartering: The model layers are divided into four quarters for the experiment, i.e., layers 0-7, 8-15, 16-23, and 24-31.

\subsection{Efficiency Comparison}
% Table generated by Excel2LaTeX from sheet 'Sheet2'
% Table generated by Excel2LaTeX from sheet 'Sheet2'
% Table generated by Excel2LaTeX from sheet 'Sheet2'
\begin{table}[htb]
  \centering
  \small
  \caption{Training Efficiency. The experiments were conducted on HiTab LLama2, using the same batch size for comparison. GPU Memory refers to the total GPU memory usage when training is stable.}
  \resizebox{0.9\columnwidth}{!}{
    \begin{tabular}{l|cc}
    \toprule
    \multicolumn{1}{c|}{Method} & Duration & GPU Memory \\
    \midrule
    Full Finetune & 45min & 280G \\
    LoRA  & 33min & 160G \\
    TableLoRA & 36min & 168G \\
    \bottomrule
    \end{tabular}%
    }
  \label{tab:efficiency}%
\end{table}%

Compared to Full Finetune, TableLoRA significantly reduces computational resources and time while bridging the performance gap between LoRA and Full Finetune. As shown in \reftab{tab:efficiency}, TableLoRA uses similar time and GPU resources as LoRA, with time being 80\% of that of Full Finetune and GPU usage being 57\%. Notably, for a fair comparison, the experiments used the same batch size. If the batch size were increased so that TableLoRA and Full Finetune used the same GPU memory, TableLoRA could achieve even better time efficiency.

\section{Prompt}
An example of the prompt is shown below (the example is the sample in \reffig{fig:error_eg}):

\lstset{  
  basicstyle=\small\ttfamily,  
  frame=none,  
  breaklines=true,
  literate=  
    {[TAB]}{{[TAB]}}5
    {[CELL]}{{[CELL]}}6 
    {[ROW]}{{[ROW]}}5
}  
  
\begin{lstlisting}
<s> [INST] This is a hierarchical table question answering task. The goal for this task is to answer the given question based on the given table. The table might be hierarchical. Here is the table to answer this question. Answer the question.
/*
[TAB][ROW][CELL]  low income[CELL]  total[CELL]  total[CELL]  canadian-born[CELL]  canadian-born[CELL]  immigrant[CELL]  immigrant[ROW][CELL]  low income[CELL]  female[CELL]  male[CELL]  female[CELL]  male[CELL]  female[CELL]  male[ROW][CELL]  low income[CELL]  percentage[CELL]  percentage[CELL]  percentage[CELL]  percentage[CELL]  percentage[CELL]  percentage[ROW][CELL]  total age groups[CELL][CELL][CELL][CELL][CELL][CELL][ROW][CELL]  visible minority[CELL]  21.9[CELL]  21.1[CELL]  19.3[CELL]  18.5[CELL]  22.0[CELL]  21.0[ROW][CELL]  not a visible minority[CELL]  14.3[CELL]  12.2[CELL]  14.2[CELL]  12.2[CELL]  14.3[CELL]  12.3[ROW][CELL]  under 15 years[CELL][CELL][CELL][CELL][CELL][CELL][ROW][CELL]  visible minority[CELL]  25.4[CELL]  25.2[CELL]  22.3[CELL]  21.8[CELL]  34.3[CELL]  36.2[ROW][CELL]  not a visible minority[CELL]  15.2[CELL]  15.2[CELL]  14.9[CELL]  14.9[CELL]  26.1[CELL]  25.7[ROW][CELL]  15 to 24 years[CELL][CELL][CELL][CELL][CELL][CELL][ROW][CELL]  visible minority[CELL]  26.3[CELL]  26.2[CELL]  18.6[CELL]  17.9[CELL]  29.2[CELL]  28.6[ROW][CELL]  not a visible minority[CELL]  15.8[CELL]  13.7[CELL]  15.4[CELL]  13.3[CELL]  20.8[CELL]  18.7[ROW][CELL]  25 to 54 years[CELL][CELL][CELL][CELL][CELL][CELL][ROW][CELL]  visible minority[CELL]  20.7[CELL]  19.3[CELL]  12.6[CELL]  11.1[CELL]  21.3[CELL]  19.8[ROW][CELL]  not a visible minority[CELL]  12.7[CELL]  11.2[CELL]  12.5[CELL]  10.9[CELL]  14.3[CELL]  13.7[ROW][CELL]  55 to 64 years[CELL][CELL][CELL][CELL][CELL][CELL][ROW][CELL]  visible minority[CELL]  17.1[CELL]  16.8[CELL]  17.3[CELL]  16.9[CELL]  17.0[CELL]  16.7[ROW][CELL]  not a visible minority[CELL]  14.4[CELL]  13.2[CELL]  14.5[CELL]  13.2[CELL]  13.4[CELL]  13.1[ROW][CELL]  65 years and over[CELL][CELL][CELL][CELL][CELL][CELL][ROW][CELL]  visible minority[CELL]  17.3[CELL]  14.3[CELL]  15.1[CELL]  9.8[CELL]  17.4[CELL]  14.4[ROW][CELL]  not a visible minority[CELL]  16.2[CELL]  9.5[CELL]  17.1[CELL]  10.0[CELL]  12.9[CELL]  None 
*/
Table Caption: The table caption is this table displays the results of prevalence of low income. the information is grouped by low income (appearing as row headers), total, canadian-born, immigrant, female and male, calculated using percentage units of measure (appearing as column headers).
Question: within the population that did not belong to a visible minority group, what was the percentage of canadian-born women aged 15 to 24 in a low-income situation?
The answer is:
[/INST]
\end{lstlisting}

\end{document}